\pdfoutput=1

\documentclass[11pt]{article}

\usepackage[]{EMNLP2022}

\usepackage{times}
\usepackage{latexsym}

\usepackage[T1]{fontenc}

\usepackage[utf8]{inputenc}

\usepackage{microtype}

\usepackage{inconsolata}
\usepackage{todonotes}
\usepackage{multirow}
\usepackage{float}
\usepackage{dblfloatfix}
\usepackage{amsmath}
\usepackage{amsfonts}

\DeclareMathOperator*{\argmax}{arg\,max}

\title{Adam Mickiewicz University at WMT 2022:\\ NER-Assisted and Quality-Aware Neural Machine Translation}

\author{Artur Nowakowski$\normalfont{\textsuperscript{ * 1,2}}$ \and Gabriela Pałka$\normalfont{\textsuperscript{ * 1,3}}$ \and Kamil Guttmann$\normalfont{\textsuperscript{ † 1,2}}$ \and Mikołaj Pokrywka $\normalfont{\textsuperscript{† 1,2}}$ \\
        $\textsuperscript{1}$ Adam Mickiewicz University, Poznań, Poland \\
        $\textsuperscript{2}$ Poleng, Poznań, Poland \\
        $\textsuperscript{3}$ Applica.ai, Warsaw, Poland \\
        \texttt{\{artur.nowakowski,gabriela.palka\}@amu.edu.pl}, \texttt{\{kamgut,mikpok1\}@st.amu.edu.pl}}

\begin{document}
\maketitle
\def\thefootnote{*}\footnotetext{AN and GP contributed equally.}\def\thefootnote{\arabic{footnote}}
\def\thefootnote{$\dagger$}\footnotetext{KG and MP contributed equally.}\def\thefootnote{\arabic{footnote}}
\begin{abstract}
This paper presents Adam Mickiewicz University's (AMU) submissions to the constrained track of the WMT 2022 General MT Task. We participated in the Ukrainian $\leftrightarrow$ Czech translation directions. The systems are a weighted ensemble of four models based on the Transformer (big) architecture. The models use source factors to utilize the information about named entities present in the input. Each of the models in the ensemble was trained using only the data provided by the shared task organizers. A noisy back-translation technique was used to augment the training corpora. One of the models in the ensemble is a document-level model, trained on parallel and synthetic longer sequences. During the sentence-level decoding process, the ensemble generated the n-best list. The n-best list was merged with the n-best list generated by a single document-level model which translated multiple sentences at a time. Finally, existing quality estimation models and minimum Bayes risk decoding were used to rerank the n-best list so that the best hypothesis was chosen according to the COMET evaluation metric. According to the automatic evaluation results, our systems rank first in both translation directions.
\end{abstract}

\section{Introduction}
We describe Adam Mickiewicz University's submissions to the constrained track of the WMT 2022 General MT Task.
We participated in the Ukrainian $\leftrightarrow$ Czech translation directions -- a low-resource translation scenario between closely related languages.

The data provided by the shared task organizers was thoroughly cleaned and filtered, as described in section~\ref{sec-data}.

The approach described in section~\ref{sec-approach} is based on combining various MT enhancement methods, including transfer learning from a high-resource language pair~\cite{transfer-learning-aji, transfer-learning-zoph}, noisy back-translation~\cite{noisy-backtranslation}, NER-assisted translation~\cite{ner-factors}, document-level translation, model ensembling, quality-aware decoding~\cite{quality-aware-decoding}, and on-the-fly domain adaptation~\cite{UnsupervisedAdaptation}.

The results leading to the final submissions are presented in section~\ref{sec-results}. Additionally, we performed a statistical significance test with paired bootstrap resampling~\cite{koehn-2004-statistical}, comparing the baseline solution with the final submission on the test set reference translations released by the shared task organizers. According to the automatic evaluation results based on COMET~\cite{comet} scores, our systems rank first in both translation directions.

\begin{table*}[t]
\centering
\begin{tabular}{llrl} \hline
\multicolumn{1}{l}{\textbf{Data type}}   &         & \textbf{Sentences}  & \textbf{Corpora} \\  \hline
\multirow{2}{*}{Monolingual cs} & available     & 448,528,116          & \multirow{2}{9cm}{News crawl, Europarl v10, News Commentary, Common Crawl, Extended Common Crawl, Leipzig Corpora} \\
                                & used & 59,999,553  \\ \hline
\multirow{2}{*}{Monolingual uk} & available     & 70,526,415          & \multirow{2}{9cm}{News crawl, UberText Corpus, Leipzig Corpora, Legal Ukrainian}  \\ 
                                & used  & 59,152,329 \\ \hline
\multirow{2}{*}{Parallel cs-uk} & available     & 12,630,806          & \multirow{2}{9cm}{Opus, WikiMatrix, ELRC -- EU acts in Ukrainian}  \\
                                & used & 8,623,440 \\ \hline
\end{tabular}
\caption{Statistics of the total available corpora and the corpora used for system training after filtering.}
\label{data}
\end{table*}

\section{Data}
\label{sec-data}
In the initial stage of system preparation, the data was cleaned and filtered using the OpusFilter~\cite{opusfilter} toolkit. With the use of the toolkit, language detection filtering based on fastText~\cite{joulin2016bag} was performed, duplicates were removed, and simple heuristics based on sentence length were applied. Then, using Moses~\cite{moses} pre-processing scripts, punctuation was normalized and non-printing characters removed. Finally, the text was tokenized into subword units using SentencePiece~\cite{sentencepiece} with the unigram language model algorithm~\cite{unigram-subword}. For Ukrainian$\rightarrow$Czech and Czech$\rightarrow$Ukrainian models trained from scratch, we used separate vocabularies for the source and the target language. Each vocabulary consisted of 32,000 units.

We used concatenated data from the Flores-101~\cite{flores} benchmark (flores101-dev, flores101-devtest) for our development set, as provided by the task organizers.

Table~\ref{data} shows statistics for the total available corpora in the constrained track and the corpora used for system training after filtering. 

\section{Approach}
\label{sec-approach}
We used the Marian~\cite{marian} toolkit for all of our experiments. Our model architecture follows the Transformer (big)~\cite{transformer} settings. For all model training, we used 4x NVIDIA A100 80GB GPUs.

\subsection{Transfer Learning}
For our initial experiments, we used transfer learning~\cite{transfer-learning-aji, transfer-learning-zoph} from the high-resource Czech$\rightarrow$English language pair. We used only the parallel data provided by the organizers to train the model in this direction. In this case, we created a single joint vocabulary for three languages (Czech, English, Ukrainian), consisting of 32,000 units. The Czech$\rightarrow$English model was fine-tuned for the Ukrainian$\rightarrow$Czech and Czech$\rightarrow$Ukrainian language directions. Our later experiments showed that there were no gains in translation quality compared with models trained from scratch using separate vocabularies for source and target languages -- the upside was that the models took less time to converge during training.

\subsection{Noisy Back-Translation}
We used models created by the transfer learning approach to produce synthetic training data through noisy back-translation~\cite{noisy-backtranslation}. Specifically, we applied Gumbel noise to the output layer and sampled from the full model distribution. We used monolingual data available in the constrained track, which included all \char`\~59M Ukrainian sentences after filtering and \char`\~60M randomly selected Czech sentences.

After training the model with concatenated parallel and back-translated corpora, we replaced the training data with filtered parallel data and further fine-tuned the model. We kept the same settings as in the first training pass, training the model until it converged on the development set.

\begin{figure*}[hbt!]
\begin{verbatim}
Hlavní|p0 inspektor|p0 organizace|p0 RSPCA|p3 pro|p0 Nový|p2 Jižní|p2 Wales|p2
David|p1 O'Shannessy|p1 televizi|p0 ABC|p5 sdělil|p0 ,|p0 že|p0 dohled|p0 nad|p0 
jatky|p0 a|p0 jejich|p0 kontroly|p0 by|p0 měly|p0 být|p0 v|p0 Austrálii|p2
samozřejmostí|p0 .|p0

_Hlavní|p0 _inspektor|p0 _organizace|p0 _R|p3 SP|p3 CA|p3 _pro|p0 _Nový|p2 _Jižní|p2
_Wales|p2 _David|p1 _O|p1 '|p1 S|p1 han|p1 ness|p1 y|p1 _televizi|p0 _A|p5 BC|p5
_sdělil|p0 ,|p0 _že|p0 _dohled|p0 _nad|p0 _ja|p0 tky|p0 _a|p0 _jejich|p0 _kontroly|p0
_by|p0 _měly|p0 _být|p0 _v|p0 _Austrálii|p2 _samozřejmost|p0 í|p0 .|p0
\end{verbatim}
\caption{An example of a sentence tagged with NER source factors before and after subword encoding.}
\label{ner-tagged-example}
\end{figure*}

\begin{table*}[hbt!]
\centering
\begin{tabular}{l|rrrr|rrrr}
\hline
\multicolumn{1}{c|}{\textbf{}} & \multicolumn{4}{c|}{\textbf{cs}}                                           & \multicolumn{4}{c}{\textbf{uk}}                                            \\ \hline
\textbf{Category}              & \textbf{train-bt} & \textbf{train-parallel} & \textbf{dev} & \textbf{test} & \textbf{train-bt} & \textbf{train-parallel} & \textbf{dev} & \textbf{test} \\ \hline
PER                            & 33,633,602        & 1,545,658               & 747          & 306           & 30,778,893        & 1,623,370               & 827          & 478           \\
LOC                            & 24,552,404        & 1,954,319               & 1,191         & 454           & 18,178,736        & 1,912,604               & 1,197         & 771           \\
ORG                            & 29,380,436        & 1,997,685               & 566          & 314           & 24,117,485        & 2,221,371               & 544          & 606           \\
MISC                           & -                 & -                       & -            & -             & 4,140,394         & 893,867                 & 168          & 76            \\
PRO                            & 5,452,326         & 1,104,860               & 172          & 59            & -                 & -                       & -            & -             \\
EVT                            & 1,150,301         & 111,563                 & 83           & 10            & -                 & -                       & -            & -             \\ \hline
\end{tabular}
\caption{The number of recognized named entity categories in the training, development and test data. The training data statistics are split into \textit{train-bt}, which was created by noisy back-translation, and \textit{train-parallel}, which is the filtered parallel training data.}
\label{ner-statistics}
\end{table*}

\subsection{NER-Assisted Translation}
\label{sec:ner}
Translation in domains such as news, social or conversational texts, and e-commerce is a specialized task, involving such challenges as localization, product names, and mentions of people or events in the content of documents. In such a case, it proved helpful to use off-the-shelf solutions for recognizing named entities. For Czech, the Slavic BERT model~\cite{slavic=-bert} was used, with which entities such as persons (PER), locations (LOC), organizations (ORG), products (PRO), and events (EVT) were tagged. Due to the lack of support for the Ukrainian language in the Slavic BERT model, the Stanza Named Entity Recognition module~\cite{stanza} was used to detect entities in the Ukrainian text, recognizing  persons (PER), locations (LOC), organizations (ORG), and miscellaneous items (MISC). With these ready-made solutions, the parallel and back-translated corpora were tagged. The named entity categories were then numbered to assign appropriate source factors to words in the text, supporting the translation process. The source factors were later transferred to subwords in a trivial way.

Source factors~\cite{factors} have previously been used to take into account various characteristics of words during the translation process. For example, morphological information, part-of-speech tags, and syntactic dependencies have been added as input to neural machine translation systems to improve the translation quality.

In the same way, it is possible to add information about named entities found in the text~\cite{ner-factors}, making it easier for the model to translate them correctly. However, the AMU machine translation system does not distinguish between inside-outside-beginning (IOB) tags~\cite{iob-tags}, treating the named entity tag names as a whole. Specifically, we introduce the following source factors:

\begin{itemize}
    \item p0: source factor denoting a normal token,
    \item p1: source factor denoting the PER category,
    \item p2: source factor denoting the LOC category,
    \item p3: source factor denoting the ORG category,
    \item p4: source factor denoting the MISC category,
    \item p5: source factor denoting the PRO category,
    \item p6: source factor denoting the EVT category.
\end{itemize}

An example of a tagged sentence is shown in Figure~\ref{ner-tagged-example}.

Models were trained in two settings: concatenation and sum. In the first setting, the factor embedding had a size of 16 and was concatenated with the token embedding. In the second setting, the factor embedding was equal to the size of the token embedding (1024) and was summed with it.

As shown in Table~\ref{dev-full-results}, we observe an increase in the string-based evaluation metrics (chrF and BLEU) while COMET scores remain about the same. This is in accordance with~\citet{amrhein2022identifying}, who show that COMET models are not sufficiently sensitive to discrepancies in named entities.

Table~\ref{ner-statistics} presents the numbers of recognized named entity categories in the training, development and test data.

\begin{figure*}[hbt!]
\begin{verbatim}
Netvrdím, že bakteriální celulóza jednou nahradí bavlnu, kůži, nebo jiné látky. 
<SEP> Ale myslím, že by to mohl být chytrý a udržitelný přírůstek k našim stále
vzácnějším přírodním zdrojům. <SEP> Možná že se nakonec tyto bakterie neuplatní
v módě, ale jinde. <SEP> Zkuste si třeba představit, že si vypěstujeme lampu,
židli, auto, nebo třeba dům. <SEP> Má otázka tedy zní: Co byste si v budoucnu
nejraději vypěstovali vy?
\end{verbatim}
\caption{An example document consisting of five sentences separated with \texttt{<SEP>} tags.}
\label{document-example}
\end{figure*}

\subsection{Document-Level Translation}

Our work on document-level translation is based on a simple data concatenation method, similar to~\citet{junczys-dowmunt-2019-microsoft} and~\citet{scherrer-etal-2019-analysing}.

As our training data, we use parallel document-level datasets (GNOME, KDE4, TED2020, QED), as well as synthetically created data, concatenating random sentences to match the desired input length.
Specifically, we merge datasets created in the following ways as a single, large dataset:

\begin{itemize}
    \item Curr $\rightarrow$ Curr: sentence-level parallel data,
    \item Prev + Curr $\rightarrow$ Prev + Curr: previous sentence given as a context,
    \item 50T $\rightarrow$ 50T: a fixed window of 50 tokens after subword encoding,
    \item 100T $\rightarrow$ 100T: a fixed window of 100 tokens after subword encoding,
    \item 250T $\rightarrow$ 250T: a fixed window of 250 tokens after subword encoding,
    \item 500T $\rightarrow$ 500T: a fixed window of 500 tokens after subword encoding.
\end{itemize}

By concatenating such datasets, we allow the model to gradually learn how to translate longer input sequences. It is also capable of sentence-level translation. To separate sentences from each other, we introduced a \verb|<SEP>| tag. An example of a document-level input sequence is shown in Figure~\ref{document-example}. All data used to train the document-level model were tagged with NER source factors, including the back-translated data.

\subsection{Weighted Ensemble}
We created a weighted ensemble of four best-performing models.
It consisted of the following models:
\begin{itemize}
    \item (A) sentence-level models trained with NER source factors (concat 16),
    \item (B) sentence-level model trained with NER source factors (sum),
    \item (C) document-level model trained with NER source factors (concat 16).
\end{itemize}
In this case, the document-level model was used only for the sentence-level translation.
The optimal weights for each model were selected using a grid search method.
For the specific language pairs, we used the following model and weight combinations:

\begin{itemize}
    \item Czech $\rightarrow$ Ukrainian: 1.0 $\cdot$ (2$\times$A) + 0.8 $\cdot$ (B) + 0.6 $\cdot$ (C),
    \item Ukrainian $\rightarrow$ Czech: 1.0 $\cdot$ (2$\times$A) + 0.8 $\cdot$ (B) + 0.4 $\cdot$ (C).
\end{itemize}

\subsection{Quality-Aware Decoding}

Having the final model ensemble, we created an n-best list containing 200 translations for each sentence with beam search. Then we merged it with a second n-best list containing 50 translations for each sentence, created by a single document-level model with document-level decoding. This enabled the use of quality-aware decoding~\cite{quality-aware-decoding}.

\begin{figure}[H]
\center
\includegraphics[width=0.48\textwidth]{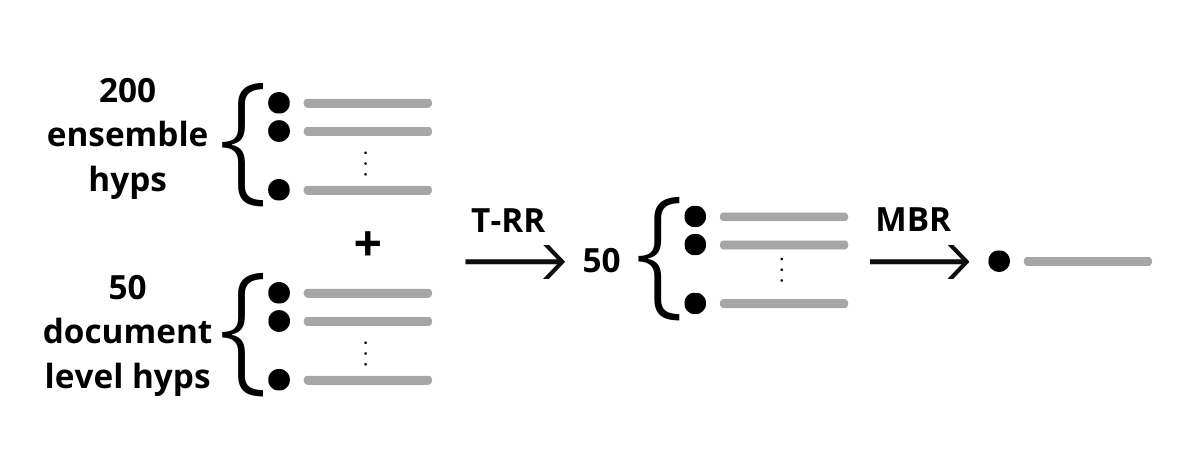}
\caption{A two-stage (T-RR $\rightarrow$ MBR) quality-aware decoding process. 200 hypotheses generated by the ensemble are merged with 50 hypotheses generated by the document-level model. A tuned reranker is used to prune the total number of hypotheses to 50, and these are then used as input for minimum Bayes risk decoding.}
\label{fig:quality-aware}
\end{figure}

We applied a two-stage quality-aware decoding mechanism: pruning hypotheses using a tuned reranker (T-RR) and minimum Bayes risk (MBR) decoding~\cite{mbr-1, mbr-2}, as shown in Figure \ref{fig:quality-aware}.

First, we tuned a reranker on the development set, using as features NMT model scores, as well as existing QE models based on TransQuest~\cite{transquest} and COMET~\cite{comet}, which are based on Direct Assessment (DA)~\cite{direct-assessment} scores or MQM~\cite{mqm} scores. Specifically, we used:

\begin{itemize}
    \item model ensemble log-likelihood $\log p_{\theta}(y|x)$ scores,
    \item TransQuest QE model trained on DA scores (\texttt{monotransquest-da-multilingual}),
    \item COMET QE model trained on MQM scores (\texttt{wmt21-comet-qe-mqm}),
    \item COMET QE model trained on DA scores (\texttt{wmt21-comet-qe-da}).
\end{itemize}

We tuned the feature weights to maximize the COMET reference-based evaluation metric value using MERT~\cite{mert}.

After tuning the reranker, we used it to prune the n-best list from 250 to 50 hypotheses per input sentence.
The resulting n-best list was used for minimum Bayes risk decoding, using the COMET reference-based metric as the utility function. 
Minimum Bayes risk decoding seeks, from the set of hypotheses, the hypothesis with the highest expected utility.

\begin{align}
\label{eq:MC-expectation}
    \hat{y}_{\textsc{mbr}} &= \argmax_{y\in\bar{\mathcal{Y}}} ~\underbrace{\mathbb{E}_{Y \sim p_\theta(y \mid x)}[ u(Y, y)]}_{\textstyle \approx ~\frac{1}{M}\sum_{j=1}^{M}u(y^{(j)}, y)}
\end{align}
Equation~\ref{eq:MC-expectation} shows that the expectation can be approximated as a Monte Carlo sum using model samples $y^{(1)}, \ldots, y^{(M)} \sim p_\theta(y|x)$.
In practice, the translation with the highest expected utility can be chosen by comparing each hypothesis $y \in \bar{\mathcal{Y}}$ with all other hypotheses in the set.

The described two-stage quality-aware decoding process allowed us to further optimize our system for the COMET evaluation metric, which has been shown to have a high correlation with human judgements~\cite{kocmi-etal-2021-ship}.

\subsection{Post-Processing}
The final step involved post-processing. We applied the following post-processing steps for each best obtained translation:

\begin{itemize}
    \item transfer of emojis from the source to the translation using word alignment based on SimAlign~\cite{jalili-sabet-etal-2020-simalign},
    \item restoration of quotation marks appropriate for a given language,
    \item restoration of capitalization (e.g. if the source sentence was fully uppercased),
    \item restoration of punctuation, exclamation and question marks (if a source sentence ends with such a mark, we make the translation do likewise),
    \item replacement of three consecutive dots with an ellipsis,
    \item restoration of bullet points and enumeration (e.g. if the source sentence starts with a number or a bullet point),
    \item deletion of consecutively repeated words.
\end{itemize}

\subsection{On-The-Fly Domain Adaptation}
The General MT Task tests the MT system's performance on multiple domains. Therefore, we investigated the possibility of improving our translation system with the on-the-fly domain adaptation method. 

This experiment was based on~\citet{UnsupervisedAdaptation}. Our idea was to retrieve similar sentences from the training data for each input sentence and to fine-tune the model on their translations. After the translation of a single sentence is complete, the model is reset to the original parameters. We used Apache Lucene~\cite{mccandless2010lucene} as our translation memory to search for similar sentences. We indexed all of the training data and used the Marian dynamic adaptation feature. We compared the translation quality with and without the retrieved context. The experiments were carried out with a different similarity score used to choose similar sentence pairs for the fine-tuning process. We empirically modified the learning rate and the number of epochs to find optimal values that improved the translation quality.

Table~\ref{runtime-adaptation-results} shows the results of the aforementioned experiments on the full development set. We found that only a small number of sentences in the training data were similar to those present in the development set. The results showed that tuning the model on similar sentences from the training data did not significantly improve translation quality. In the end, we decided not to use this method in our WMT 2022 submission.


\begin{table}[ht]
\centering
\begin{tabular}{lrrr} \hline
                        \textbf{Approach} & \textbf{Sim. score} & \textbf{COMET}  & \textbf{chrF}  \\ \hline
\multirow{1}{1.5cm}{Baseline} & - & 0.8322  & 0.5263 \\ \hline
\multirow{1}{1.5cm}{Default}  & 0.4 & 0.8316 & 0.5260 \\
\multirow{1}{1.5cm}{Best-334}  & 0.19 & 0.8322 & 0.5259 \\
\multirow{1}{1.5cm}{Best-133}  & 0.25 & 0.8323 & 0.5262 \\  \hline
\end{tabular}
\caption{Results of the on-the-fly adaptation method on the development set. The \textit{default} approach is based on~\citet{UnsupervisedAdaptation}. However, only 11 sentence pairs were found in this scenario. The experiments denoted as \textit{best-334} and \textit{best-133} used the learning rate values of 0.002 and 10 epochs. In our development set containing 2009 sentence pairs, 334 matching sentences were found in \textit{best-334} and 133 in \textit{best-133}.}
\label{runtime-adaptation-results}
\end{table}

\begin{table*}[ht]
\centering
\begin{tabular}{llrrr|rrr}
\hline
\multicolumn{2}{c}{\multirow{2}{*}{\textbf{System}}}                  & \multicolumn{3}{c|}{\textbf{uk$\rightarrow$cs}}                                                                         & \multicolumn{3}{c}{\textbf{cs$\rightarrow$uk}}                                                                         \\ \cline{3-8} 
\multicolumn{2}{c}{}                                                  & \multicolumn{1}{c}{\textbf{COMET}} & \multicolumn{1}{c}{\textbf{chrF}} & \multicolumn{1}{c|}{\textbf{BLEU}} & \multicolumn{1}{c}{\textbf{COMET}} & \multicolumn{1}{c}{\textbf{chrF}} & \multicolumn{1}{c}{\textbf{BLEU}} \\ \hline
\multicolumn{2}{l}{Baseline (transformer-big)}                        & 0.8622                             & 0.5229                            & 24.29                              & 0.7818                             & 0.5175                            & 22.64                             \\
\multicolumn{2}{l}{+back-translation}                                 & 0.9053                             & 0.5309                            & 25.41                              & 0.8356                             & 0.5280                            & 23.14                             \\
\multicolumn{1}{l|}{\multirow{2}{*}{+ner}}          & concat 16       & 0.9003                             & 0.5314                            & 25.62                              & 0.8362                             & 0.5309                            & 24.28                             \\
\multicolumn{1}{l|}{}                               & sum             & 0.8991                             & 0.5323                            & 25.87                              & 0.8421                             & 0.5302                            & 23.91                             \\
\multicolumn{1}{l|}{\multirow{2}{*}{+fine-tune}}    & concat 16       & 0.9021                             & 0.5344                            & 25.94                              & 0.8387                             & 0.5330                            & 24.51                             \\
\multicolumn{1}{l|}{}                               & sum             & 0.8990                             & 0.5357                            & 25.99                              & 0.8456                             & 0.5321                            & 24.24                             \\
\multicolumn{2}{l}{+ensemble}                                         & 0.9066                             & 0.5376                            & \textbf{26.36}                     & 0.8522                             & 0.5373                            & \textbf{24.85}                    \\
\multicolumn{2}{l}{+quality-aware}                                    & 0.9874                             & 0.5376                            & 25.42                              & 0.9238                             & 0.5384                            & 24.50                             \\
\multicolumn{2}{l}{+post-processing}                                  & \textbf{0.9883}                    & \textbf{0.5392}                   & 25.89                              & \textbf{0.9240}                    & \textbf{0.5388}                   & 24.63                             \\ \hline
\multicolumn{1}{c}{\multirow{2}{*}{Document-level}} & sent-level dec. & 0.8942                             & 0.5326                            & 25.47                              & 0.8350                             & 0.5289                            & 23.92                             \\
\multicolumn{1}{c}{}                                & doc-level dec.  & 0.8920                             & 0.5324                            & 25.44                              & 0.8356                             & 0.5297                            & 23.78                             \\ \hline
\end{tabular}
\caption{Results of COMET, chrF and BLEU automatic evaluation metrics on the concatenated datasets flores101-dev and flores-101-devtest. ChrF and BLEU metrics were computed with sacreBLEU. Document-level model evaluation includes added back-translation, NER source factors (concat 16) and fine-tuning.}
\label{dev-full-results}
\end{table*}

\begin{table*}[ht]
\centering
\begin{tabular}{lcrr|crr}
\hline
\multicolumn{1}{c}{\multirow{2}{*}{\textbf{System}}} & \multicolumn{3}{c|}{\textbf{uk$\rightarrow$cs}}                                                     & \multicolumn{3}{c}{\textbf{cs$\rightarrow$uk}}                                                              \\ \cline{2-7} 
\multicolumn{1}{c}{}                                 & \textbf{COMET}             & \multicolumn{1}{c}{\textbf{chrF}} & \multicolumn{1}{c|}{\textbf{BLEU}} & \textbf{COMET}             & \multicolumn{1}{c}{\textbf{chrF}} & \multicolumn{1}{c}{\textbf{BLEU}} \\ \hline
Baseline (transformer-big)                          & \multicolumn{1}{r}{0.8315} & 0.5627 & 31.79                            & \multicolumn{1}{r}{0.8008} & 0.5849                            & 31.43                          \\
Final submission                                     & \multicolumn{1}{r}{\textbf{1.0488}}      & \textbf{0.6066}                                 & \textbf{37.03}                              & \multicolumn{1}{r}{\textbf{0.9944}}      & \textbf{0.6153}                                 & \textbf{34.74}                         \\ \hline
\end{tabular}
\caption{Results of COMET, chrF and BLEU automatic evaluation metrics on the test set. ChrF and BLEU metrics were computed with sacreBLEU. The final submission results are statistically significant (\textit{p} < 0.05).}
\label{test-full-results}
\end{table*}

\section{Results}
\label{sec-results}
The results of our experiments are presented in Table~\ref{dev-full-results}. We evaluated our models with the COMET\footnote{COMET scores were computed with the \texttt{wmt20-comet-da} model.}~\cite{comet}, chrF~\cite{popovic-2015-chrf} and BLEU~\cite{papineni-etal-2002-bleu} automatic evaluation metrics. ChrF and BLEU scores were computed with the sacreBLEU\footnote{
BLEU signature: nrefs:1|case:mixed|eff:no|tok:13a\\|smooth:exp|version:2.0.0}\footnote{chrF signature: nrefs:1|case:mixed|eff:yes|nc:6|nw:0\\|space:no|version:2.0.0}~\cite{post-2018-call} tool. We also include scores for the document-level model. In this case, the scores include improvements added by back-translation, NER source factors and fine-tuning. The document-level evaluation was split into sentence-level decoding and document-level decoding. In the first scenario, the model translates a single sentence at a time, which is not different from a sentence-level model. In the second scenario, the model translates concatenated chunks of at most 250 subword tokens at a time. 

We found that the largest gain in the COMET value was achieved due to the quality-aware decoding method, at the cost of BLEU value. The chrF value remained the same in the Ukrainian$\rightarrow$Czech translation direction, while it increased slightly in the Czech$\rightarrow$Ukrainian direction. As discussed in section~\ref{sec:ner}, the inclusion of NER source factors helped the model with the translation of named entities, which is not well reflected in the COMET value, as this metric is not sufficiently sensitive to discrepancies in named entities~\cite{amrhein2022identifying}.

Table~\ref{test-full-results} shows results for our final submissions compared with the baseline. 
We performed a statistical significance test with paired bootstrap resampling~\cite{koehn-2004-statistical}, running 1000 resampling trials to confirm that our submissions are statistically significant (\textit{p} < 0.05).

\section{Conclusions}
We describe Adam Mickiewicz University's (AMU) submissions to the WMT 2022 General MT Task in the Ukrainian $\leftrightarrow$ Czech translation directions. Our experiments cover a range of MT enhancement methods, including transfer learning, back-translation, NER-assisted translation, document-level translation, weighted ensembling, quality-aware decoding, and on-the-fly domain adaptation. We found that using a combination of these methods on the test set leads to a +0.22 (26.13\%) increase in COMET scores in the Ukrainian$\rightarrow$Czech translation direction and a +0.19 (24.18\%) increase in the Czech$\rightarrow$Ukrainian direction, compared with the baseline model. According to the COMET automatic evaluation results, our systems rank first in both translation directions.

\bibliography{anthology,custom}

\begin{thebibliography}{35}
\expandafter\ifx\csname natexlab\endcsname\relax\def\natexlab#1{#1}\fi

\bibitem[{Aji et~al.(2020)Aji, Bogoychev, Heafield, and
  Sennrich}]{transfer-learning-aji}
Alham~Fikri Aji, Nikolay Bogoychev, Kenneth Heafield, and Rico Sennrich. 2020.
\newblock \href {https://doi.org/10.18653/v1/2020.acl-main.688} {In neural
  machine translation, what does transfer learning transfer?}
\newblock In \emph{Proceedings of the 58th Annual Meeting of the Association
  for Computational Linguistics}, pages 7701--7710, Online. Association for
  Computational Linguistics.

\bibitem[{Amrhein and Sennrich(2022)}]{amrhein2022identifying}
Chantal Amrhein and Rico Sennrich. 2022.
\newblock Identifying weaknesses in machine translation metrics through minimum
  {B}ayes risk decoding: A case study for {COMET}.
\newblock \emph{arXiv preprint arXiv:2202.05148}.

\bibitem[{Arkhipov et~al.(2019)Arkhipov, Trofimova, Kuratov, and
  Sorokin}]{slavic=-bert}
Mikhail Arkhipov, Maria Trofimova, Yuri Kuratov, and Alexey Sorokin. 2019.
\newblock \href {https://doi.org/10.18653/v1/W19-3712} {Tuning multilingual
  transformers for language-specific named entity recognition}.
\newblock In \emph{Proceedings of the 7th Workshop on Balto-Slavic Natural
  Language Processing}, pages 89--93, Florence, Italy. Association for
  Computational Linguistics.

\bibitem[{Aulamo et~al.(2020)Aulamo, Virpioja, and Tiedemann}]{opusfilter}
Mikko Aulamo, Sami Virpioja, and J{\"o}rg Tiedemann. 2020.
\newblock \href {https://doi.org/10.18653/v1/2020.acl-demos.20}
  {{O}pus{F}ilter: A configurable parallel corpus filtering toolbox}.
\newblock In \emph{Proceedings of the 58th Annual Meeting of the Association
  for Computational Linguistics: System Demonstrations}, pages 150--156.
  Association for Computational Linguistics.

\bibitem[{Edunov et~al.(2018)Edunov, Ott, Auli, and
  Grangier}]{noisy-backtranslation}
Sergey Edunov, Myle Ott, Michael Auli, and David Grangier. 2018.
\newblock \href {https://doi.org/10.18653/v1/D18-1045} {Understanding
  back-translation at scale}.
\newblock In \emph{Proceedings of the 2018 Conference on Empirical Methods in
  Natural Language Processing}, pages 489--500, Brussels, Belgium. Association
  for Computational Linguistics.

\bibitem[{Farajian et~al.(2017)Farajian, Turchi, Negri, and
  Federico}]{UnsupervisedAdaptation}
M.~Amin Farajian, Marco Turchi, Matteo Negri, and Marcello Federico. 2017.
\newblock \href {https://doi.org/10.18653/v1/W17-4713} {Multi-domain neural
  machine translation through unsupervised adaptation}.
\newblock In \emph{Proceedings of the Second Conference on Machine
  Translation}, pages 127--137, Copenhagen, Denmark. Association for
  Computational Linguistics.

\bibitem[{Fernandes et~al.(2022)Fernandes, Farinhas, Rei, De~Souza, Ogayo,
  Neubig, and Martins}]{quality-aware-decoding}
Patrick Fernandes, Ant{\'o}nio Farinhas, Ricardo Rei, Jos{\'e} De~Souza, Perez
  Ogayo, Graham Neubig, and Andre Martins. 2022.
\newblock \href {https://aclanthology.org/2022.naacl-main.100} {Quality-aware
  decoding for neural machine translation}.
\newblock In \emph{Proceedings of the 2022 Conference of the North American
  Chapter of the Association for Computational Linguistics: Human Language
  Technologies}, pages 1396--1412, Seattle, United States. Association for
  Computational Linguistics.

\bibitem[{Goyal et~al.(2022)Goyal, Gao, Chaudhary, Chen, Wenzek, Ju, Krishnan,
  Ranzato, Guzm{\'a}n, and Fan}]{flores}
Naman Goyal, Cynthia Gao, Vishrav Chaudhary, Peng-Jen Chen, Guillaume Wenzek,
  Da~Ju, Sanjana Krishnan, Marc{'}Aurelio Ranzato, Francisco Guzm{\'a}n, and
  Angela Fan. 2022.
\newblock \href {https://doi.org/10.1162/tacl_a_00474} {The {F}lores-101
  evaluation benchmark for low-resource and multilingual machine translation}.
\newblock \emph{Transactions of the Association for Computational Linguistics},
  10:522--538.

\bibitem[{Graham et~al.(2013)Graham, Baldwin, Moffat, and
  Zobel}]{direct-assessment}
Yvette Graham, Timothy Baldwin, Alistair Moffat, and Justin Zobel. 2013.
\newblock \href {https://aclanthology.org/W13-2305} {Continuous measurement
  scales in human evaluation of machine translation}.
\newblock In \emph{Proceedings of the 7th Linguistic Annotation Workshop and
  Interoperability with Discourse}, pages 33--41, Sofia, Bulgaria. Association
  for Computational Linguistics.

\bibitem[{Jalili~Sabet et~al.(2020)Jalili~Sabet, Dufter, Yvon, and
  Sch{\"u}tze}]{jalili-sabet-etal-2020-simalign}
Masoud Jalili~Sabet, Philipp Dufter, Fran{\c{c}}ois Yvon, and Hinrich
  Sch{\"u}tze. 2020.
\newblock \href {https://doi.org/10.18653/v1/2020.findings-emnlp.147}
  {{S}im{A}lign: High quality word alignments without parallel training data
  using static and contextualized embeddings}.
\newblock In \emph{Findings of the Association for Computational Linguistics:
  EMNLP 2020}, pages 1627--1643, Online. Association for Computational
  Linguistics.

\bibitem[{Joulin et~al.(2016)Joulin, Grave, Bojanowski, and
  Mikolov}]{joulin2016bag}
Armand Joulin, Edouard Grave, Piotr Bojanowski, and Tomas Mikolov. 2016.
\newblock Bag of tricks for efficient text classification.
\newblock \emph{arXiv preprint arXiv:1607.01759}.

\bibitem[{Junczys-Dowmunt(2019)}]{junczys-dowmunt-2019-microsoft}
Marcin Junczys-Dowmunt. 2019.
\newblock \href {https://doi.org/10.18653/v1/W19-5321} {{M}icrosoft translator
  at {WMT} 2019: Towards large-scale document-level neural machine
  translation}.
\newblock In \emph{Proceedings of the Fourth Conference on Machine Translation
  (Volume 2: Shared Task Papers, Day 1)}, pages 225--233, Florence, Italy.
  Association for Computational Linguistics.

\bibitem[{Junczys-Dowmunt et~al.(2018)Junczys-Dowmunt, Grundkiewicz, Dwojak,
  Hoang, Heafield, Neckermann, Seide, Germann, Aji, Bogoychev, Martins, and
  Birch}]{marian}
Marcin Junczys-Dowmunt, Roman Grundkiewicz, Tomasz Dwojak, Hieu Hoang, Kenneth
  Heafield, Tom Neckermann, Frank Seide, Ulrich Germann, Alham~Fikri Aji,
  Nikolay Bogoychev, Andr{\'e} F.~T. Martins, and Alexandra Birch. 2018.
\newblock \href {https://doi.org/10.18653/v1/P18-4020} {{M}arian: Fast neural
  machine translation in {C}++}.
\newblock In \emph{Proceedings of {ACL} 2018, System Demonstrations}, pages
  116--121, Melbourne, Australia. Association for Computational Linguistics.

\bibitem[{Kocmi et~al.(2021)Kocmi, Federmann, Grundkiewicz, Junczys-Dowmunt,
  Matsushita, and Menezes}]{kocmi-etal-2021-ship}
Tom Kocmi, Christian Federmann, Roman Grundkiewicz, Marcin Junczys-Dowmunt,
  Hitokazu Matsushita, and Arul Menezes. 2021.
\newblock \href {https://aclanthology.org/2021.wmt-1.57} {To ship or not to
  ship: An extensive evaluation of automatic metrics for machine translation}.
\newblock In \emph{Proceedings of the Sixth Conference on Machine Translation},
  pages 478--494, Online. Association for Computational Linguistics.

\bibitem[{Koehn(2004)}]{koehn-2004-statistical}
Philipp Koehn. 2004.
\newblock \href {https://aclanthology.org/W04-3250} {Statistical significance
  tests for machine translation evaluation}.
\newblock In \emph{Proceedings of the 2004 Conference on Empirical Methods in
  Natural Language Processing}, pages 388--395, Barcelona, Spain. Association
  for Computational Linguistics.

\bibitem[{Koehn et~al.(2007)Koehn, Hoang, Birch, Callison-Burch, Federico,
  Bertoldi, Cowan, Shen, Moran, Zens, Dyer, Bojar, Constantin, and
  Herbst}]{moses}
Philipp Koehn, Hieu Hoang, Alexandra Birch, Chris Callison-Burch, Marcello
  Federico, Nicola Bertoldi, Brooke Cowan, Wade Shen, Christine Moran, Richard
  Zens, Chris Dyer, Ond{\v{r}}ej Bojar, Alexandra Constantin, and Evan Herbst.
  2007.
\newblock \href {https://aclanthology.org/P07-2045} {{M}oses: Open source
  toolkit for statistical machine translation}.
\newblock In \emph{Proceedings of the 45th Annual Meeting of the Association
  for Computational Linguistics Companion Volume Proceedings of the Demo and
  Poster Sessions}, pages 177--180, Prague, Czech Republic. Association for
  Computational Linguistics.

\bibitem[{Kudo(2018)}]{unigram-subword}
Taku Kudo. 2018.
\newblock \href {https://doi.org/10.18653/v1/P18-1007} {Subword regularization:
  Improving neural network translation models with multiple subword
  candidates}.
\newblock In \emph{Proceedings of the 56th Annual Meeting of the Association
  for Computational Linguistics (Volume 1: Long Papers)}, pages 66--75,
  Melbourne, Australia. Association for Computational Linguistics.

\bibitem[{Kudo and Richardson(2018)}]{sentencepiece}
Taku Kudo and John Richardson. 2018.
\newblock \href {https://doi.org/10.18653/v1/D18-2012} {{S}entence{P}iece: A
  simple and language independent subword tokenizer and detokenizer for neural
  text processing}.
\newblock In \emph{Proceedings of the 2018 Conference on Empirical Methods in
  Natural Language Processing: System Demonstrations}, pages 66--71, Brussels,
  Belgium. Association for Computational Linguistics.

\bibitem[{Kumar and Byrne(2002)}]{mbr-1}
Shankar Kumar and William Byrne. 2002.
\newblock \href {https://doi.org/10.3115/1118693.1118712} {Minimum {B}ayes-risk
  word alignments of bilingual texts}.
\newblock In \emph{Proceedings of the 2002 Conference on Empirical Methods in
  Natural Language Processing ({EMNLP} 2002)}, pages 140--147. Association for
  Computational Linguistics.

\bibitem[{Kumar and Byrne(2004)}]{mbr-2}
Shankar Kumar and William Byrne. 2004.
\newblock \href {https://aclanthology.org/N04-1022} {Minimum {B}ayes-risk
  decoding for statistical machine translation}.
\newblock In \emph{Proceedings of the Human Language Technology Conference of
  the North {A}merican Chapter of the Association for Computational
  Linguistics: {HLT}-{NAACL} 2004}, pages 169--176, Boston, Massachusetts, USA.
  Association for Computational Linguistics.

\bibitem[{Lommel et~al.(2014)Lommel, Uszkoreit, and Burchardt}]{mqm}
Arle Lommel, Hans Uszkoreit, and Aljoscha Burchardt. 2014.
\newblock Multidimensional quality metrics (mqm): A framework for declaring and
  describing translation quality metrics.
\newblock \emph{Revista Tradum{\`a}tica: tecnologies de la traducci{\'o}},
  (12):455--463.

\bibitem[{McCandless et~al.(2010)McCandless, Hatcher, and
  Gospodneti{\'c}}]{mccandless2010lucene}
M.~McCandless, E.~Hatcher, and O.~Gospodneti{\'c}. 2010.
\newblock \href {https://books.google.it/books?id=XrJBPgAACAAJ} {\emph{Lucene
  in Action}}.
\newblock Manning Pubs Co Series. Manning.

\bibitem[{Modrzejewski et~al.(2020)Modrzejewski, Exel, Buschbeck, Ha, and
  Waibel}]{ner-factors}
Maciej Modrzejewski, Miriam Exel, Bianka Buschbeck, Thanh-Le Ha, and Alexander
  Waibel. 2020.
\newblock \href {https://aclanthology.org/2020.eamt-1.6} {Incorporating
  external annotation to improve named entity translation in {NMT}}.
\newblock In \emph{Proceedings of the 22nd Annual Conference of the European
  Association for Machine Translation}, pages 45--51, Lisboa, Portugal.
  European Association for Machine Translation.

\bibitem[{Och(2003)}]{mert}
Franz~Josef Och. 2003.
\newblock \href {https://doi.org/10.3115/1075096.1075117} {Minimum error rate
  training in statistical machine translation}.
\newblock In \emph{Proceedings of the 41st Annual Meeting on Association for
  Computational Linguistics -- Volume 1}, ACL '03, page 160–167, USA.
  Association for Computational Linguistics.

\bibitem[{Papineni et~al.(2002)Papineni, Roukos, Ward, and
  Zhu}]{papineni-etal-2002-bleu}
Kishore Papineni, Salim Roukos, Todd Ward, and Wei-Jing Zhu. 2002.
\newblock \href {https://doi.org/10.3115/1073083.1073135} {{B}leu: a method for
  automatic evaluation of machine translation}.
\newblock In \emph{Proceedings of the 40th Annual Meeting of the Association
  for Computational Linguistics}, pages 311--318, Philadelphia, Pennsylvania,
  USA. Association for Computational Linguistics.

\bibitem[{Popovi{\'c}(2015)}]{popovic-2015-chrf}
Maja Popovi{\'c}. 2015.
\newblock \href {https://doi.org/10.18653/v1/W15-3049} {chr{F}: character
  n-gram {F}-score for automatic {MT} evaluation}.
\newblock In \emph{Proceedings of the Tenth Workshop on Statistical Machine
  Translation}, pages 392--395, Lisbon, Portugal. Association for Computational
  Linguistics.

\bibitem[{Post(2018)}]{post-2018-call}
Matt Post. 2018.
\newblock \href {https://doi.org/10.18653/v1/W18-6319} {A call for clarity in
  reporting {BLEU} scores}.
\newblock In \emph{Proceedings of the Third Conference on Machine Translation:
  Research Papers}, pages 186--191, Brussels, Belgium. Association for
  Computational Linguistics.

\bibitem[{Qi et~al.(2020)Qi, Zhang, Zhang, Bolton, and Manning}]{stanza}
Peng Qi, Yuhao Zhang, Yuhui Zhang, Jason Bolton, and Christopher~D. Manning.
  2020.
\newblock \href {https://doi.org/10.18653/v1/2020.acl-demos.14} {{S}tanza: A
  {P}ython natural language processing toolkit for many human languages}.
\newblock In \emph{Proceedings of the 58th Annual Meeting of the Association
  for Computational Linguistics: System Demonstrations}, pages 101--108,
  Online. Association for Computational Linguistics.

\bibitem[{Ramshaw and Marcus(1995)}]{iob-tags}
Lance Ramshaw and Mitch Marcus. 1995.
\newblock \href {https://aclanthology.org/W95-0107} {Text chunking using
  transformation-based learning}.
\newblock In \emph{Third Workshop on Very Large Corpora}.

\bibitem[{Ranasinghe et~al.(2020)Ranasinghe, Orasan, and Mitkov}]{transquest}
Tharindu Ranasinghe, Constantin Orasan, and Ruslan Mitkov. 2020.
\newblock \href {https://doi.org/10.18653/v1/2020.coling-main.445}
  {{T}rans{Q}uest: Translation quality estimation with cross-lingual
  transformers}.
\newblock In \emph{Proceedings of the 28th International Conference on
  Computational Linguistics}, pages 5070--5081, Barcelona, Spain (Online).
  International Committee on Computational Linguistics.

\bibitem[{Rei et~al.(2020)Rei, Stewart, Farinha, and Lavie}]{comet}
Ricardo Rei, Craig Stewart, Ana~C Farinha, and Alon Lavie. 2020.
\newblock \href {https://doi.org/10.18653/v1/2020.emnlp-main.213} {{COMET}: A
  neural framework for {MT} evaluation}.
\newblock In \emph{Proceedings of the 2020 Conference on Empirical Methods in
  Natural Language Processing (EMNLP)}, pages 2685--2702, Online. Association
  for Computational Linguistics.

\bibitem[{Scherrer et~al.(2019)Scherrer, Tiedemann, and
  Lo{\'a}iciga}]{scherrer-etal-2019-analysing}
Yves Scherrer, J{\"o}rg Tiedemann, and Sharid Lo{\'a}iciga. 2019.
\newblock \href {https://doi.org/10.18653/v1/D19-6506} {Analysing concatenation
  approaches to document-level {NMT} in two different domains}.
\newblock In \emph{Proceedings of the Fourth Workshop on Discourse in Machine
  Translation (DiscoMT 2019)}, pages 51--61, Hong Kong, China. Association for
  Computational Linguistics.

\bibitem[{Sennrich and Haddow(2016)}]{factors}
Rico Sennrich and Barry Haddow. 2016.
\newblock \href {https://doi.org/10.18653/v1/W16-2209} {Linguistic input
  features improve neural machine translation}.
\newblock In \emph{Proceedings of the First Conference on Machine Translation:
  Volume 1, Research Papers}, pages 83--91, Berlin, Germany. Association for
  Computational Linguistics.

\bibitem[{Vaswani et~al.(2017)Vaswani, Shazeer, Parmar, Uszkoreit, Jones,
  Gomez, Kaiser, and Polosukhin}]{transformer}
Ashish Vaswani, Noam Shazeer, Niki Parmar, Jakob Uszkoreit, Llion Jones,
  Aidan~N. Gomez, {\L}ukasz Kaiser, and Illia Polosukhin. 2017.
\newblock \href
  {https://proceedings.neurips.cc/paper/2017/file/3f5ee243547dee91fbd053c1c4a845aa-Paper.pdf}
  {Attention is all you need}.
\newblock In \emph{Advances in Neural Information Processing Systems},
  volume~30. Curran Associates, Inc.

\bibitem[{Zoph et~al.(2016)Zoph, Yuret, May, and
  Knight}]{transfer-learning-zoph}
Barret Zoph, Deniz Yuret, Jonathan May, and Kevin Knight. 2016.
\newblock \href {https://doi.org/10.18653/v1/D16-1163} {Transfer learning for
  low-resource neural machine translation}.
\newblock In \emph{Proceedings of the 2016 Conference on Empirical Methods in
  Natural Language Processing}, pages 1568--1575, Austin, Texas. Association
  for Computational Linguistics.

\end{thebibliography}
\bibliographystyle{acl_natbib}

\end{document}